\documentclass[11pt,a4paper]{article}

\usepackage[utf8]{inputenc}
\usepackage[T1]{fontenc}
\usepackage{amsmath,amssymb}
\usepackage{graphicx}
\usepackage{booktabs}
\usepackage{tabularx}
\usepackage{longtable}
\usepackage{hyperref}
\usepackage[margin=1in]{geometry}
\usepackage[authoryear,round]{natbib}
\usepackage{enumitem}
\usepackage{array}
\usepackage{multirow}
\usepackage{caption}
\usepackage{float}
\usepackage{xr-hyper}
\usepackage{siunitx}

\hypersetup{
    colorlinks=true,
    linkcolor=blue,
    citecolor=blue,
    urlcolor=blue
}

\title{HEBATRON: A Hebrew-Specialized Open-Weight Mixture-of-Experts Language Model\\[0.5em]\large Technical Report}

\author{
 \textbf{Noam Kayzer}\thanks{These authors contributed equally to this work.} \and
 \textbf{Dan Revital}\footnotemark[1] \and
 \textbf{Ori Bar Joseph}\footnotemark[1] \and
Smadar Arvatz \and
Or Levi \and
Tal Geva \and
Shaltiel Shmidman \and
Amir DN Cohen \and
Noam Ordan \and
Omer Baruch \and
Kate Zinkovskaia \and
Zevi Apini \and
 \textbf{Sarel Weinberger}\thanks{Corresponding author}
\\
PwC Next
}
\date{April 26, 2026}

\begin{document}

\maketitle

\begin{abstract}
We present \textsc{Hebatron}, a Hebrew-specialized open-weight large language model built on the NVIDIA Nemotron-3 sparse Mixture-of-Experts architecture. Training employs a three-phase easy-to-hard curriculum with continuous anti-forgetting anchoring, followed by supervised fine-tuning on 2 million bilingual Hebrew--English samples. The curriculum ordering alone yields a 3-point aggregate benchmark gain over the reversed configuration. \textsc{Hebatron} achieves a Hebrew reasoning average of 73.8\%, outperforming the best open source models in Hebrew and remaining competitive with Gemma-3-27B-IT on GSM8K-HE and Israeli Trivia, while activating only ${\sim}3$B parameters per forward pass across a 30B-parameter model, delivering approximately $9{\times}$ higher inference throughput at native context lengths up to 65,536 tokens. To our knowledge, this is the first language-specific adaptation of the Nemotron-3 architecture for any target language, and the first open-weight Hebrew-specialized MoE model with native long-context support. Model weights are released openly to support further research in Hebrew and Semitic-language NLP.
\end{abstract}

\section{Introduction}

The rapid advancement of large language models (LLMs) has transformed the landscape of natural language processing. These models now enable human-level performance across a broad range of reasoning, generation, and comprehension tasks \citep{openai2024gpt4o, gemini2024, anthropic2024claude3}. However, the overwhelming majority of frontier model development remains concentrated on English-centric training regimes Non-English languages particularly morphologically complex, low-resource languages receiving substantially less representation in both pretraining corpora and post-training alignment pipelines \citep{touvron2023llama2, joshi2020state}. This imbalance has resulted in a persistent and well-documented performance gap between English and non-English speakers in their access to capable, culturally grounded AI systems.

Hebrew is a particularly compelling and challenging target for language model localization. As a Semitic language with rich morphological structure, non-concatenative templatic derivation, and widespread orthographic ambiguity arising from the optional use of diacritics (niqqud), Hebrew imposes demands on language model architecture and pretraining data that differ fundamentally from those of Indo-European languages \citep{chriqui2022hebert, shmidman2024dictalm2}. Words in Hebrew are formed via root-and-pattern morphology, where the same three- or four-consonant root can surface in dozens of morphologically distinct forms depending on binyan and context. This templatic structure, combined with the prevalence of prefixed prepositions, conjunctions, and definiteness markers that attach to word stems, creates acute challenges for tokenization, lemmatization, and downstream semantic understanding that standard multilingual tokenizers and models handle poorly \citep{antoun2020arabert, chriqui2022hebert}. Beyond morphology, Hebrew is a right-to-left language embedded in a predominantly left-to-right digital ecosystem, and the corpus of high-quality digitized Hebrew text spanning literature, law, journalism, and academic discourse is orders of magnitude smaller than the English web. These properties jointly make Hebrew a low-resource language in the practical sense relevant to large-scale pretraining, despite its status as a living language with millions of native speakers and a vibrant digital presence.

Prior work across Arabic NLP 
\citep{antoun2020arabert, inoue2021camelbert}, Indic languages \citep{kakwani2020indicnlpsuite, doddapaneni2023indic}, and East Asian languages \citep{cui2023chinese, fujii2024continual} has demonstrated that general-purpose multilingual models consistently underperform language-specific models on tasks requiring deep cultural knowledge, morphological precision, and pragmatic reasoning. The emerging paradigm of sovereign language model development has shown considerable promise as a scalable strategy for closing this gap without incurring the full compute cost of training from scratch. Most recently, this paradigm has produced state-of-the-art results in two closely related linguistic domains: DictaLM-3.0 \citep{shmidman2026dictalm3}, which establishes the current frontier for open-weight Hebrew-capable models through large-scale continued pretraining on 130B Hebrew tokens; and Command-R7B-Arabic \citep{cohere2025arabic}, which demonstrates that a compact model localized on Arabic, a morphologically kindred Semitic language can match or exceed substantially larger general-purpose models on language-specific benchmarks. Both cases reinforce the central finding of domain-adaptive pretraining research \citep{gururangan2020dont, howard2018ulmfit}: continued pretraining on high-quality, in-domain data yields robust improvements across downstream tasks, provided the distribution is carefully designed to preserve cross-lingual reasoning parity \citep{pfeiffer-etal-2020-mad, ustun2024aya}.

A critical dimension of effective localization is the selection of the foundation model from which continued pretraining is initialized. We build upon the NVIDIA Nemotron-3-Nano-30B-A3B-Base-BF16 model \citep{nvidia2025nemotron}, a sparse Mixture-of-Experts (MoE) architecture trained on a large-scale, publicly documented pretraining corpus. Crucially, the public availability of Nemotron's pretraining data distributions enables a strategic reintegration of original source data, particularly high-quality English reasoning corpora during the localization process. This transparency serves as a vital anti-forgetting anchor: by co-training on a curated subset of the foundation model's original training signal, we mitigate catastrophic forgetting of baseline reasoning and English-language proficiency, a failure mode consistently observed in naive continued pretraining pipelines \citep{goodfellow2013catastrophic, kirkpatrick2017overcoming, luo2023catastrophic}. The scalable efficiency of the MoE architecture further ensures that the 30B-parameter model maintains strong compute-to-performance characteristics throughout the localization process \citep{fedus2021switch, jiang2024mixtral}.

Beyond base model selection, the training corpus design follows principles established in large-scale multilingual pretraining initiatives including BLOOM \citep{leScao2022bloom}, MADLAD-400 \citep{kudugunta2023madlad400}, and Aya \citep{ustun2024aya}, where curated, linguistically balanced data mixtures were shown to be decisive factors in downstream language quality and reasoning generalization. For Semitic-language modeling specifically, the most recent and instructive precedents are DictaLM-3.0 \citep{shmidman2026dictalm3} for Hebrew and Command-R7B-Arabic \citep{cohere2025arabic} for Arabic both demonstrating that dedicated, high-quality pretraining corpora tailored to capture templatic morphology, orthographic variation, and culturally grounded world knowledge are prerequisite to achieving competitive performance in morphologically rich, low-resource languages. FineWeb2 \citep{penedo2024fineweb}, the most comprehensive high-quality multilingual web crawl available at training time, forms the primary web-sourced Hebrew component of our corpus.

The sequential structure of our pretraining curriculum is motivated by a growing body of evidence on data ordering in LLM pretraining. \citet{bengio2009curriculum} established the foundational principle that presenting training examples from easier to harder accelerates learning and improves generalization. Recent work validates this at LLM scale: \citet{chen2023skillit} demonstrate that easy-to-hard curriculum ordering consistently accelerates convergence and yields sustained downstream improvements. In our Hebrew localization setting, this motivates initializing pretraining on formally structured, morphologically regular sources such as literary texts, legal documents, parliamentary protocols, and academic corpora before exposing the model to the noisier patterns of social media and informal web content. Context length extension is handled as a third sequential stage, following established progressive scaling practices \citep{peng2023yarn, chen2024longlora}, employing a dedicated long-document corpus filtered to documents exceeding 2,000 words, enabling coherent multi-document reasoning at up to 65,536 tokens.

Following the Continuous Pre-training (CPT) stage, instruction tuning via supervised fine-tuning (SFT) on a bilingual Hebrew-English mixture of 2M high-quality samples aligns the localized base model for instruction-following and multi-step reasoning \citep{ouyang2022instructgpt, wang2022selfinstruct, wei2022cot}.

The primary contributions of this work are as follows:

\begin{itemize}

\item A Hebrew-specialized large language model \citep{hebatron2026} based on the Nemotron-3-Nano-30B-A3B-Base-BF16 MoE 
architecture \citep{nvidia2025nemotron}, trained through a structured three-phase curriculum pretraining pipeline on 
approximately 154B tokens of Hebrew and English data, supporting native context lengths of 
up to 65,536 tokens.

\item \textbf{Novel Architecture-Language Localization:} To our knowledge, this work represents the first documented instance of a complete localization pipeline—integrating both large-scale continuous pre-training (CPT) and supervised fine-tuning (SFT)—for any target language utilizing the NVIDIA Nemotron-3 sparse Mixture-of-Experts (MoE) architecture \citep{nvidia2025nemotron}.

\item \textbf{Curriculum-ordered localization strategy:} We empirically validate that an 
easy-to-hard data ordering --- formal sources before colloquial and social media content --- 
yields superior morphological quality and benchmark performance compared to the reverse 
ordering, with an aggregate Hebrew benchmark improvement of 3.01 points (68.00 vs.\ 64.99).

\item \textbf{Large-scale multi-domain Hebrew pretraining corpus:} Spanning web, literary, 
legal, governmental, academic, news and social media sources across three curriculum phases
totaling approximately 154B tokens.

\item \textbf{Dedicated Hebrew alignment corpus:} 2M instruction-tuning samples, incorporating 
localized knowledge distillation from English reasoning pipelines alongside a procedurally 
generated Hebrew IFEval dataset targeting morphological precision.

\item \textbf{Comprehensive evaluation:} Our model achieves a Hebrew reasoning average of 
73.8\%, surpassing DictaLM-3.0-24B-Thinking (68.9\%) across automated benchmarks and 
outperforming it decisively in human preference evaluation (68.8\% of decisive votes). It 
remains competitive with Gemma-3-27B-IT on factuality and culturally grounded Hebrew 
knowledge, including Israeli Trivia (72.1\% vs.\ 70.4\%) and GSM8K~(HE) (83.3\% vs.\ 
82.8\%), while operating at approximately one-ninth the active inference compute. English 
reasoning fidelity is preserved, with an English average of 86.0\%.
Notably, the model achieves approximately \textbf{9× faster} inference compared to both Gemma-3-27B-IT and DictaLM-3.0-24B-Thinking, while operating at roughly one-ninth of the active inference compute, highlighting its efficiency advantages.

\item \textbf{Open-weight model release:} All model weights are released openly under a 
permissive license, making this the first open-weight Hebrew-specialized MoE language model 
with native 65k-token context support, providing the research community with a reproducible 
foundation for further Hebrew NLP development.

\end{itemize}

\section*{Related Work}
A detailed review of prior research encompassing language-specific adaptation, the evolution of Semitic-language modeling, and recent advances in training efficiency is provided in the Supplementary Materials. See Supplementary Materials for this context.

\section{Methods}

\subsection { Data }
\subsubsection{ Continuous Pre-training (CPT)}

\subsubsection* {Phase 1 - High-Quality Localization Seed}

\textbf{Web / General} content accounts for \textbf{36.85\%} of the training weight, serving as the primary anchor for Hebrew linguistic fluency through broad, high-quality general-domain data. This foundation is supplemented by a curated ensemble of \textbf{Cultural \& Academic}, \textbf{Legal \& Government}, and \textbf{News \& Media} sources to ensure deep domain grounding and stylistic diversity. To preserve cross-lingual reasoning fidelity, a significant English component comprising the \textbf{Nemotron} corpus was integrated into the mix, providing the necessary \textbf{STEM \& Reasoning} scaffolding to support the model's analytical capabilities from the outset. The full token distribution is detailed 
in Table~\ref{tab:phase1_corpus}.

\begin{table}[h]
\centering
\caption{Phase 1 Consolidated Multilingual Corpus: Static Token Distribution by Language and Category}
\label{tab:phase1_corpus}
\begin{tabular}{llcc}
\toprule
\textbf{Language} & \textbf{Category} & \textbf{Total Corpus Tokens} & \textbf{Corpus Composition} \\ 
\midrule
\textbf{English} & STEM \& Reasoning    & 13.60B & 45.9\%\\ 
\midrule
\textbf{Hebrew}  & Web / General        & 10.90B & 36.8\%\\
                 & Cultural \& Academic & 3.7B& 12.5\%\\
                 & Legal \& Government  & 1.10B  & 3.7\%\\
                 & News \& Media        & 0.30B  & 1.0\%\\ 
\midrule
\textbf{Total Pool} & \textbf{HE + EN}  & \textbf{29.6B}& \textbf{100.0\%}\\ 
\bottomrule
\end{tabular}
\end{table}
\begin{figure}[H]
    \centering
    \includegraphics[width=0.7\linewidth]{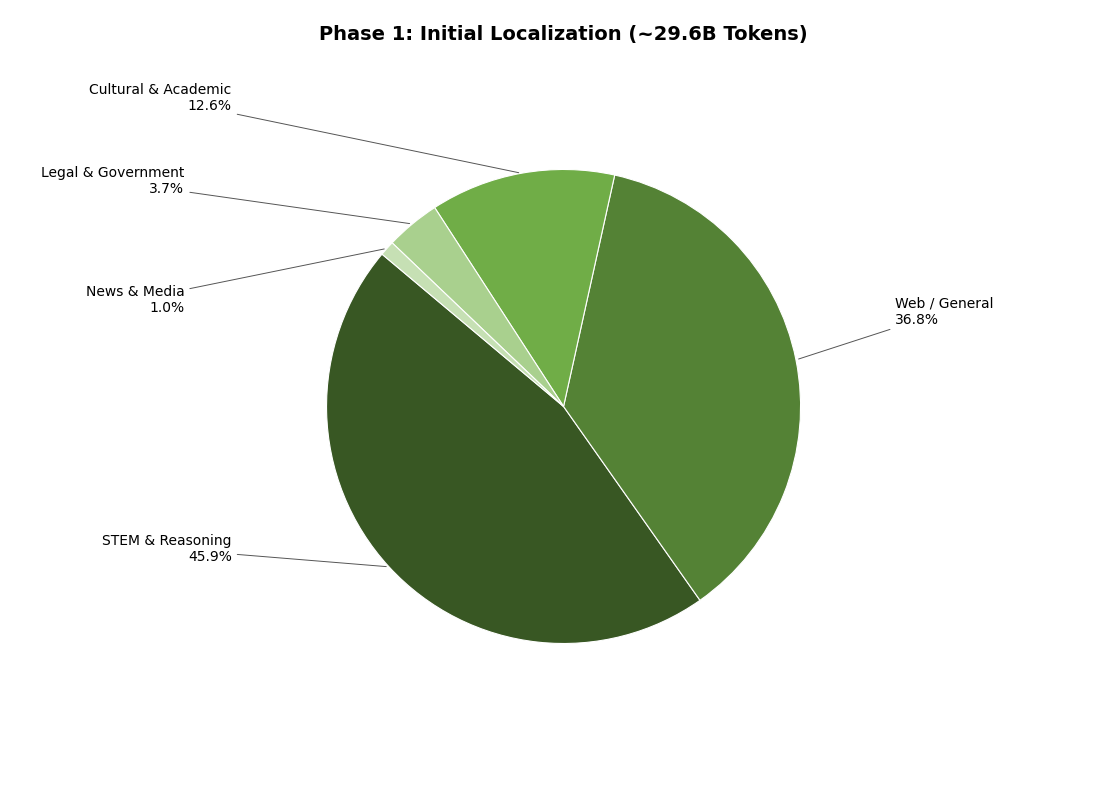}
    \caption{Data mixture of Phase 1.}
    \label{fig:data_mixture_phase1}
\end{figure}

\subsubsection*{Phase 2 - Colloquial and Broad-Domain Expansion}

The Hebrew component of Phase 2 constitutes approximately \textbf{68.5\%} of the total token pool, reflecting the phase's core objective of deepening colloquial and broad-domain coverage. \textbf{News \& Social Media} forms the largest slice at 25.93B tokens (27.2\%), covering the full register spectrum from formal journalism to informal user-generated content and serving as the main source of contemporary Hebrew usage and named-entity grounding. The \textbf{Web} component (22.16B tokens, 23.3\%) provides broad everyday lexical coverage, while \textbf{Cultural \& Academic} sources (14.71B tokens, 15.5\%) preserve the formal register grounding established in Phase 1, ensuring that exposure to noisier data does not degrade morphological precision. Smaller but purposeful contributions from \textbf{Legal \& Government} (1.27B tokens, 1.3\%) and a dedicated \textbf{Social \& Colloquial} slice (1.14B tokens, 1.2\%) maintain syntactic stability in structured Hebrew while providing focused exposure to slang and non-standard morphological forms that are largely absent from Phase 1. The full token distribution is summarized in Table~\ref{tab:phase2_aligned}.

On the English side, the \textbf{Nemotron} corpus and \textbf{Semantic Scholar} together account for \textbf{23.97\%} of the training weight, anchoring the model's multi-step reasoning and specialized academic capabilities. This is further supported by \textbf{FineWeb-Edu} \citep{penedo2024fineweb}, which provides high-quality educational content. In total, the English STEM and reasoning suite comprises approximately \textbf{31.5\%} of the final token distribution, ensuring the preservation of reasoning fidelity and technical proficiency consistent with cross-lingual adaptation frameworks \citep{conneau2020xlmr}.

\begin{table}[h]
\centering
\caption{Phase 2 Consolidated Multilingual Corpus: Static Token Distribution Aligned with Reasoning Scaling Visualization}
\label{tab:phase2_aligned}
\begin{tabular}{llcc}
\toprule
\textbf{Language} & \textbf{Category} & \textbf{Total Corpus Tokens} & \textbf{Corpus Composition} \\ 
\midrule
\textbf{English} & STEM \& Reasoning    & 29.89B & 31.5\%\\ 
\midrule
\textbf{Hebrew}  & News \& Social Media & 25.93B& 27.2\%\\
                 & Web                  & 22.16B & 23.3\%\\
                 & Cultural \& Academic & 14.71B& 15.5\%\\
                 & Social \& Colloquial & 1.14B  & 1.2\%  \\
                 & Legal \& Government  & 1.27B& 1.3\%\\ 
\midrule
\textbf{Total Pool} & \textbf{HE + EN}  & \textbf{~95.10B}& \textbf{100.0\%} \\ 
\bottomrule
\end{tabular}
\end{table}

\begin{figure}[H]
    \centering
    \includegraphics[width=0.7\linewidth]{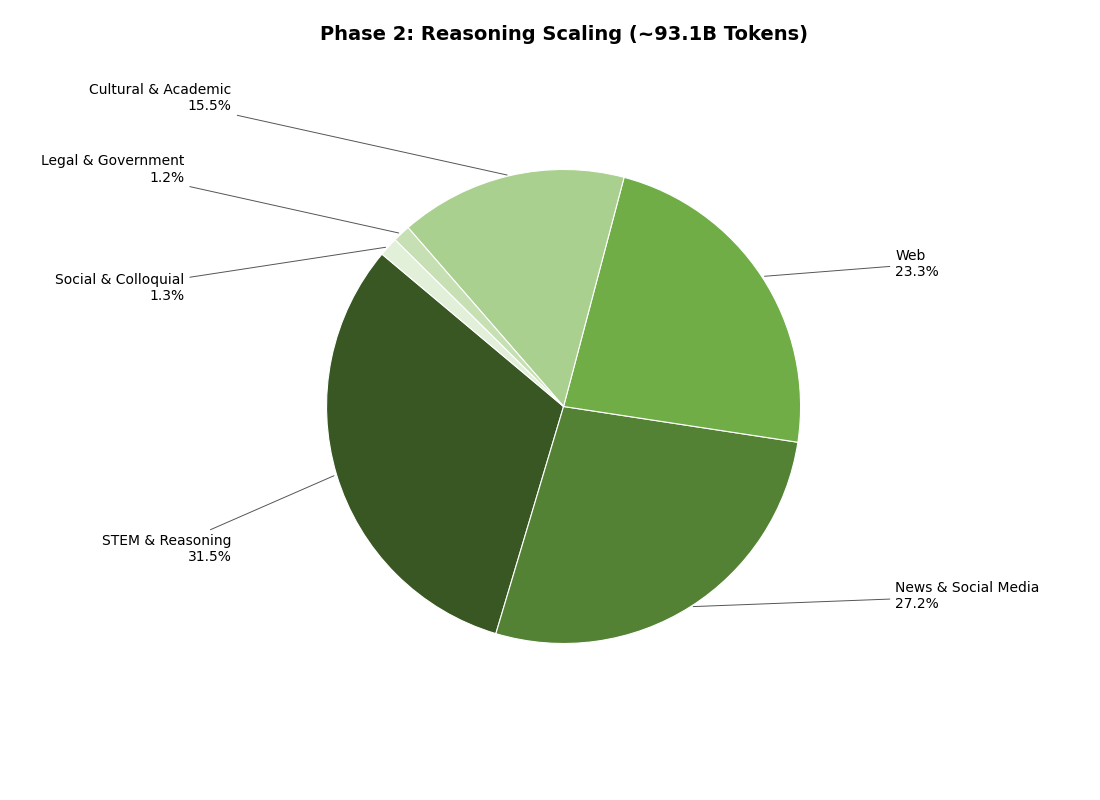}
    \caption{Data mixture of Phase 2.}
    \label{fig:data_mixture_phase2}
\end{figure}

\subsubsection*{Phase 3 - Long-Context Extension}

The training for this phase was executed on a filtered corpus of  \textbf{20.4B tokens} (14.2B Hebrew, 6.3B English), with the full data mixture detailed in 
Table~\ref{tab:phase3_aggregated}. This selection process maintains a Hebrew-dominant ratio of \textbf{69.4\% Hebrew} and \textbf{30.6\% English}—reflecting the fact that long-form Hebrew documents (legal rulings, parliamentary protocols, literary archives, and academic corpora) are naturally more prevalent in the filtered distribution. On the English side, long-document sequences were sourced from the \textbf{Nemotron corpora} \citep{nvidia2025nemotron}  to preserve long-context reasoning fidelity and ensure cross-lingual stability throughout the extension process.
\begin{table}[H]
\centering
\caption{Aggregated Phase 3 CPT Data Mixture (Context Extension) Total \texttildelow20.4B Tokens}
\label{tab:phase3_aggregated}
\small
\begin{tabular}{llcc}
\toprule
\textbf{Language} & \textbf{Category} & \textbf{Tokens} & \textbf{Mix Weight} \\
\midrule
\textbf{English} & STEM \& Reasoning    & 6.256B  & 30.63\% \\
\midrule
\textbf{Hebrew}  & Web                  & 7.038B  & 34.46\% \\
                 & Cultural \& Academic & 5.396B  & 26.42\% \\
                 & Legal \& Government  & 1.010B  & 4.95\%  \\
                 & News \& Social Media & 0.707B  & 3.46\%  \\
                 & News \& Media        & 0.012B  & 0.05\%  \\
\midrule
\textbf{Total}   & \textbf{HE + EN}     & \textbf{20.419B} & \textbf{100.0\%}\\
\bottomrule
\end{tabular}
\end{table}

\begin{figure}[H]
    \centering
    \includegraphics[width=0.7\linewidth]{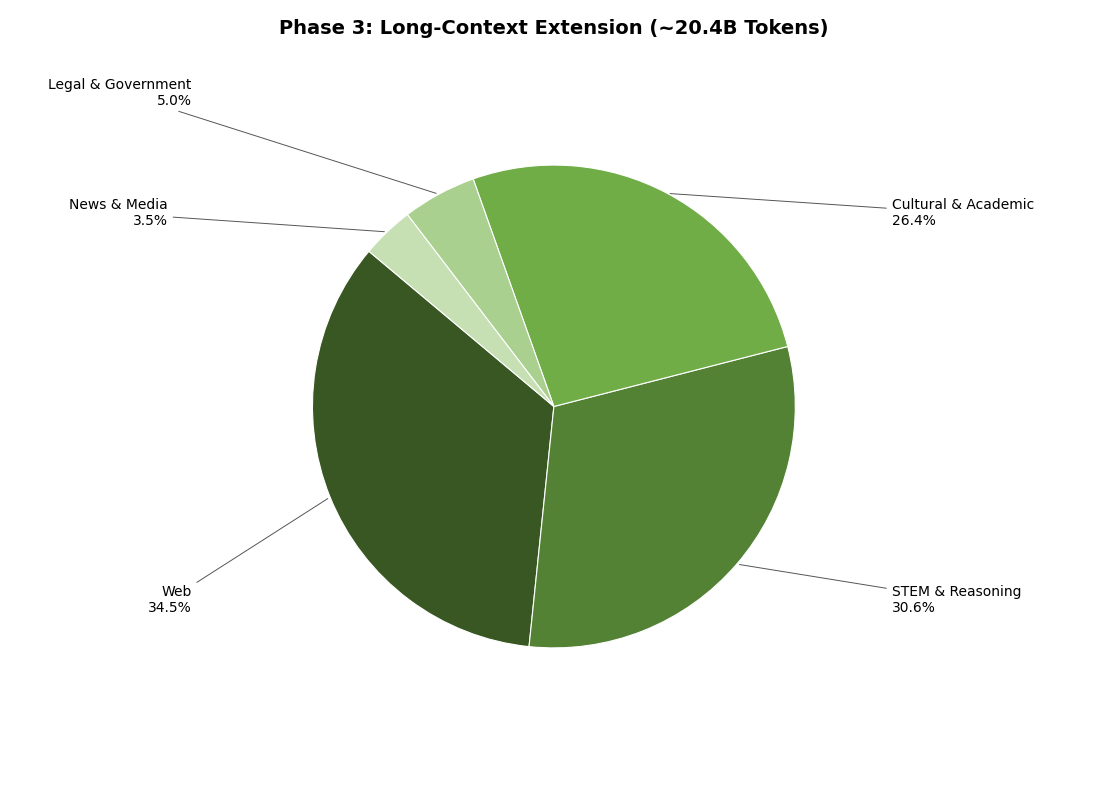}
    \caption{Data mixture of Phase 3.}
    \label{fig:data_mixture_phase3}
\end{figure}

\subsubsection{Supervised Fine-Tuning (SFT)}
\label{sec:sft-training}
The SFT corpus consists of 2M high-fidelity samples spanning seven categories, combining 
localized knowledge distillation from English reasoning pipelines, a dedicated Hebrew 
linguistic alignment dataset, and broad conversational and multi-turn coverage. The full dataset composition is summarized in Table~\ref{tab:sft_mixture}.
\subsubsection*{Localized Knowledge Distillation}
To leverage advanced reasoning traces from English-centric corpora, we implement a context-aware localization pipeline using high-capability teacher models. Training was conducted by incorporating original English source datasets in their entirety alongside corresponding localized Hebrew translations to maximize cross-lingual transfer. Language-adaptive fine-tuning (LAFT) studies suggest that localized alignment via language-specific adapters significantly enhances downstream reasoning and cross-lingual transfer \citep{pfeiffer-etal-2020-mad}.
Subsets include:

\begin{itemize}
    \item \textbf{Instruction Following:} A subset of the chat\_if collection \citep{nvidia2025nemotron} containing 678,776 total samples (combined English and Hebrew pairs).
    \item \textbf{Structured Outputs:} 9,936 samples emphasizing syntactic validity and JSON schema constraints \citep{nvidia2025nemotron}.
    \item \textbf{STEM \& Science Reasoning:} 214,358 samples sourced from Nemotron-Science-v1 \citep{nvidia2025nemotron}.
    \item \textbf{Long-Context Understanding:} 43,222 samples from the ChatQA2 collection \citep{xu2024chatqa2}, ensuring stability under extended context windows.
\end{itemize}

\subsubsection*{Hebrew IFEval (Linguistic Constraint Alignment)}
To improve native linguistic adherence, we introduced a specialized Hebrew IFEval dataset featuring 200,147 procedurally generated samples. This corpus targets essential language-specific capabilities, including morphological precision (such as the correct application of verbal patterns and morphological paradigms), prefix management, and strict adherence to complex instructional constraints.

\subsubsection*{Independent Synthetic Bitext}
To enhance cross-lingual performance and technical reasoning, we generated 187,268 synthetic SFT samples. This bilingual dataset addresses the lack of formal and professional content in standard parallel corpora like CCMatrix \citep{schwenk2021ccmatrix}, which often underrepresent specialized domains \citep{ustun2022denoising}. By using synthetic generation, we created high-quality training pairs that ensure the model can accurately understand and follow complex instructions across both languages \citep{wang2022selfinstruct}.

\subsubsection*{Conversational and Reasoning Augmentation}

We incorporated the Hermes-3 collection (663,245 samples) \citep{teknium2024hermes3} to broaden conversational and multi-turn coverage.

\begin{table}[H]
\centering
\caption{SFT Dataset Distribution and Composition}
\label{tab:sft_mixture}
\small
\begin{tabular}{llc}
\toprule
\textbf{Dataset Category} & \textbf{Lang} & \textbf{Samples (Tokens)} \\
\midrule
Instruction Following        & EN/HE & 678,776 (1.038B) \\
Conversational \& Reasoning  & EN/HE & 663,245 (434M)   \\
STEM \& Science Reasoning    & EN/HE & 214,358 (541M)   \\
Hebrew IFEval                & HE    & 200,147 (85M)    \\
Independent Synthetic Bitext & EN/HE & 187,268 (55M)    \\
Long-Context Understanding   & EN/HE & 43,222 (567M)    \\
Structured Outputs           & EN/HE & 9,936 (41M)      \\
Additional SFT Mixture       & EN/HE & 66,581 (39.4M)   \\
\midrule
\textbf{Total}               & \textbf{EN/HE} & \textbf{2,063,533 (2.8B)} \\
\bottomrule
\end{tabular}
\end{table}

\begin{figure}[H]
    \centering
    \includegraphics[width=0.7\linewidth]{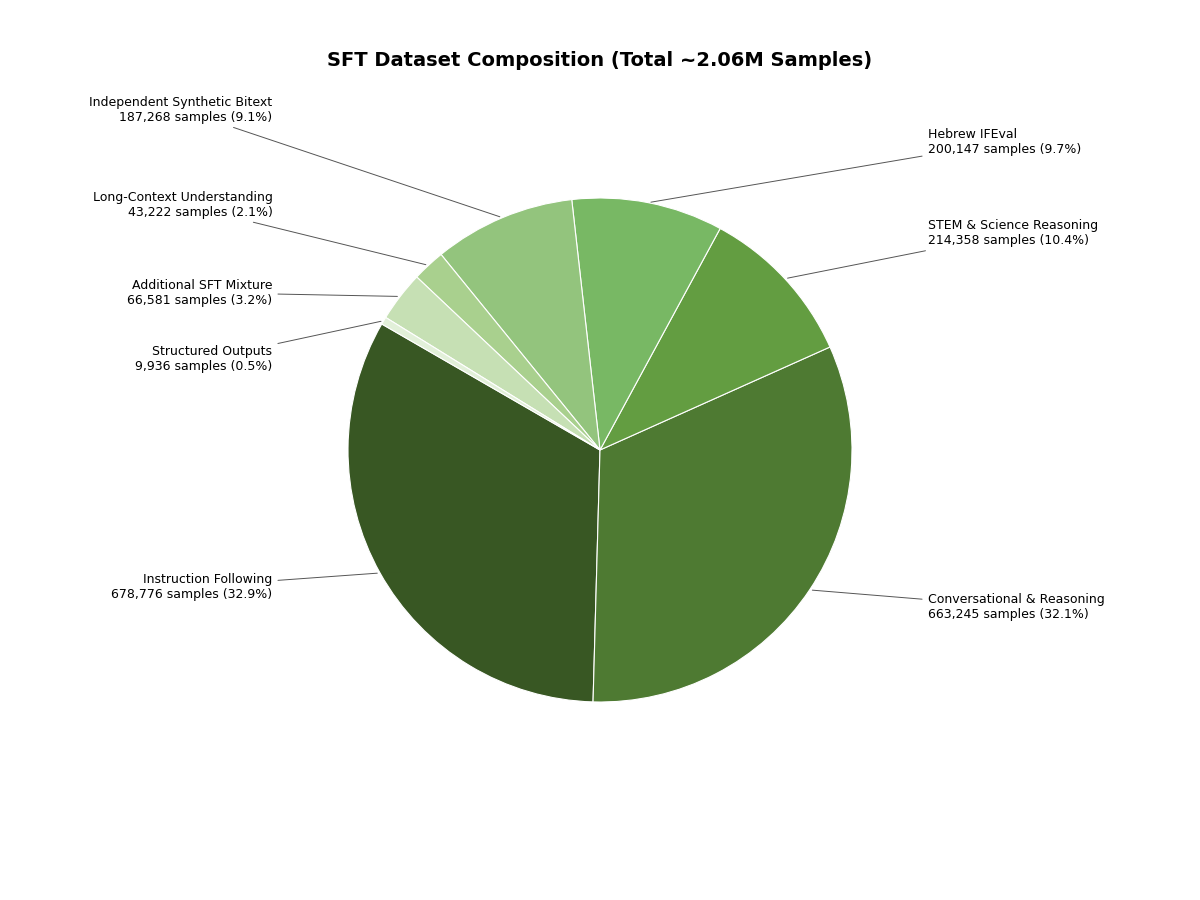}
    \caption{Distribution of supervised fine-tuning (SFT) data across 2M high-fidelity samples.}
    \label{fig:Distribution of supervised fine-tuning (SFT) data across 2M high-fidelity samples}
\end{figure}

\subsection { Data Preprocessing Pipeline  }

Prior to training, all corpus sources were passed through a structured four-stage preprocessing pipeline designed to maximize data quality while preserving semantic richness. The Gopher project \citep{rae2021gopher} established rigorous quality filtering as a cornerstone of large-scale pretraining, showing that carefully curated corpora consistently outperform larger but noisier counterparts. FineWeb \citep{penedo2024fineweb} similarly demonstrated that systematic, reproducible preprocessing pipelines combining heuristic filtering, deduplication, and quality scoring at web scale yield substantially higher-quality training distributions than generic web crawls.

\paragraph{Regex-Based Cleaning.} The first stage applies rule-based transformations to remove systematic artifacts commonly found in raw web data, including HTML and XML tags, URLs, email addresses, control characters, repeated punctuation patterns, and boilerplate scraping fragments. Rules are designed conservatively to preserve Hebrew-specific orthographic conventions and avoid overly aggressive normalization.

\paragraph{Heuristic Content Filtering.} The second stage removes documents unlikely to provide useful learning signals based on document-level heuristic criteria: documents outside predefined length bounds, those exhibiting abnormal character distributions (e.g., excessive symbols or digits), or those containing high repetition indicative of low-information content. Filtering thresholds were determined empirically \citep{penedo2024fineweb, raffel2020t5}.

\paragraph{MinHash-Based Deduplication.} The third stage mitigates duplicate and near-duplicate content using a MinHash-based deduplication strategy. Each document is represented as a set of character n-gram shingles, from which compact MinHash signatures are computed and indexed via locality-sensitive hashing (LSH) to efficiently identify similar documents. This approach scales well to large datasets while capturing both exact and approximate duplicates \citep{lee2022deduplicating, tirumala2023d4}.

\paragraph{Whitespace Normalization.} The fourth stage applies targeted whitespace normalization exclusively to web-scraped subsets, which are particularly prone to spacing inconsistencies arising from HTML parsing artifacts and encoding issues. Normalization is performed using dicta-il/dictabert-char-spacefix, a pretrained character-level model specifically designed for restoring proper spacing in Hebrew text.

\subsection { Training Methodology  }

\subsubsection{Continuous Pre-training (CPT) }
The Continuous Pre-training (CPT) phase adapts the base model to the Hebrew linguistic domain while preserving general reasoning capabilities. Continued pretraining has been shown to be an effective mechanism for domain and language adaptation without catastrophic degradation of previously acquired knowledge \citep{gururangan2020dont}. Domain-adaptive pretraining has long been shown to improve transfer performance across downstream tasks \citep{howard2018ulmfit}. Language-specific large language model initiatives across multiple regions have similarly demonstrated that continued pretraining on culturally aligned corpora improves localized reasoning capabilities \citep{fujii2024continual, huang-etal-2024-acegpt}. Adapter-based multilingual adaptation studies further demonstrate that language-specific specialization can be achieved while preserving shared representations across languages \citep{pfeiffer-etal-2020-mad}. Multilingual studies such as XLM-R \citep{conneau2020xlmr} and mT5 \citep{xue2021mt5} demonstrate that balanced multilingual mixtures enable strong cross-lingual transfer while maintaining reasoning stability, motivating our bilingual Hebrew-English training distribution. Crucially, the selection of the NVIDIA Nemotron family as our base model was driven by the public availability of its pre-training data distributions. This transparency allowed for the strategic reintegration of original source data during the localization process, serving as a vital anchor to prevent catastrophic forgetting of the model's baseline reasoning and English-language proficiency.

The CPT corpus was organized across three sequential data phases, each designed to progressively deepen linguistic and reasoning capabilities. The data selection strategy prioritized high-signal and professionally curated sources to stabilize optimization during continued pretraining. The corpus mixture follows principles established in large-scale multilingual training efforts such as BLOOM \citep{leScao2022bloom} and MADLAD-400 \citep{kudugunta2023madlad400}, where linguistic diversity and curated sampling strategies were shown to improve both language fidelity and reasoning generalization across domains. Recent multilingual model studies highlight that domain-balanced mixtures reduce catastrophic forgetting while improving downstream reasoning performance \citep{ibrahim2024simple}.

Modern advancements in Semitic-language modeling, exemplified by DictaLM-3.0 \citep{shmidman2026dictalm3} and the Arabic capabilities of Command-R7B-Arabic \citep{cohere2025arabic}, reinforce the principle that morphology-rich languages require dedicated, high-quality pretraining corpora to master templatic morphology and orthographic variation. This ensures the model captures the structural nuances of morphology and syntax required for high-register Hebrew discourse.

\subsubsection* {Phase 1 - High-Quality Localization Seed (Steps 0--4,500)}

The first phase ran for \textbf{4,500 steps} with a context length of \textbf{8,192}. This training was preformed on approximately \textbf{75.5B tokens}, which were \textbf{randomly sampled} from our curated dataset to ensure a representative distribution. Together all 3 phases of our training follow a curriculum learning strategy, structuring training from easier, well-formed material toward progressively harder and noisier data. Recent work has demonstrated that easy-to-hard curriculum ordering consistently accelerates convergence and yields sustained downstream improvements compared to random data shuffling \citep{zhang2025preference, elgaar2026curriculum}. In our localization setting, structured literary, academic, legal, and journalistic sources constitute the easy end of the curriculum: they exhibit consistent morphology, standard orthography, and well-formed syntax, properties that allow the model to internalize the formal rules of Hebrew efficiently before encountering deviations from them. We experimented with both orderings and found that the curriculum approach (easy-to-hard) yielded meaningfully better Hebrew linguistic quality and benchmark performance. In the context of Hebrew localization, we find that inverting this sequence is significantly less effective. Our primary curriculum configuration achieved an \textbf{aggregate performance of 68.00} across all benchmarks. In comparison, an evaluation of a \textbf{reversed curriculum configuration} yielded a significantly lower result, with an \textbf{average of 64.99}. This contrast underscores the effectiveness of our specific sequencing strategy in optimizing the model's learning trajectory.Despite an identical compute budget, these results further validate the efficacy of the proposed easy-to-hard curriculum for specialized linguistic adaptation.

\subsubsection*{Phase 2 - Colloquial and Broad-Domain Expansion (Steps 4,500--4,700)}

Following the structured seed phase, we implemented a targeted second stage comprising \textbf{3.36 billion tokens} (200 steps). This phase introduced increased linguistic complexity by incorporating diverse, colloquial, and social-media-derived Hebrew datasets. By shifting toward less structured, real-world data, this stage served as the ``difficult'' tier of our curriculum learning strategy, enhancing the model's robustness in informal contexts. In curriculum learning terms, tweets, forum posts, and informal web content represent significantly harder training material and are typographically noisy, morphologically inconsistent, rich in slang and abbreviations, and structurally disordered relative to the formal sources of Phase 1 \citep{zhang2025preference, elgaar2026curriculum}. Exposing the model to this register only after formal Hebrew structure was already internalized ensured that colloquial patterns were learned as extensions of a well-formed linguistic foundation rather than as competing noise. As validated in our ablations, the reverse ordering produced degraded morphological consistency and lower benchmark scores, confirming the benefit of the easy-to-hard curriculum.

The full corpus was employed in this phase, incorporating a large-scale social media crawl constructed to capture naturally occurring Hebrew-language references to prominent public figures and named entities as they appear in tweets, forum posts, and online discussions, alongside large news sources, Hebrew Tweets, and the full OSCAR multilingual web-crawl subset \citep{ortizsuarez2019oscar}. The brevity of this phase (200 steps versus 4,500 in Phase 1) reflects its targeted objective: calibrating colloquial register and broadening lexical coverage without disturbing the formal linguistic foundations established in Phase 1.

\subsubsection*{Phase 3 - Long-Context Extension}

The training for this phase was executed on \textbf{2.35 billion} tokens, which were sampled at the document level from the total filtered corpus of \textbf{20.4 billion} tokens (\num{14.2}B Hebrew, \num{6.3}B English). This subset was randomly selected to maintain the underlying distribution of the broader dataset while optimizing for computational efficiency.
To support the context extension stage, a dedicated long-document corpus was constructed by filtering the source datasets to retain only high-density, cohesive documents. This filtering strategy ensures that training sequences are semantically rich and naturally extended, which is critical for stabilizing attention mechanisms at extreme context lengths (\textbf{65,536}). By prioritizing documents with significant intrinsic length, we avoid the degenerate behavior and attention dilution that typically arise when multiple short-form documents are artificially padded or concatenated to fill long context windows.

\subsubsection*{Training Hyperparameters}
The final hyperparameters for the CPT phase are summarized in Table~\ref{tab:cpt_hyper}
The final hyperparameters for the CPT phase are summarized in Table 5. Global batch size selection was determined iteratively based on large-scale language model training scaling principles and optimizer noise considerations. Following the empirical observations of \cite{mccandlish2018empirical}, we aimed to keep the total number of optimizer updates within the stable large-scale training regime (approximately 25k–100k optimizer steps for a 250B-token CPT stage), while maintaining sufficiently large token batches to reduce gradient stochasticity. Given a context length of 8,192 tokens, the target token budget per optimizer step was derived from:

\begin{equation}
GB_{tokens} = \frac{D}{N_{steps}}
\end{equation}

where $D$ is the total CPT token budget and $N_{steps}$ is the desired number of optimizer updates. This process led iteratively to a global batch configuration of 2,048 sequences per step, corresponding to approximately 16.7M tokens per optimizer update.

Furthermore, the learning-rate configuration followed the noise-preserving scaling formulation proposed by \cite{smith2017bayesian}, where the effective optimization noise approximately scales as $\eta/\sqrt{B}$, with $\eta$ denoting the learning rate and $B$ the effective global batch size. In the final CPT configuration, we used a conservative peak learning rate of $5 \times 10^{-5}$ and a minimum learning rate of $5 \times 10^{-6}$, together with warmup and decay scheduling. This choice deliberately avoids aggressive linear LR scaling when moving to a large global batch of 2,048 sequences, prioritizing stable continued pretraining over maximum update magnitude.

\begin{table}[H]
\centering
\caption{Continuous Pre-training (CPT) Hyperparameters}
\label{tab:cpt_hyper}
\small
\begin{tabular}{lll}
\toprule
\textbf{Hyperparameter} & \textbf{Stage 1 \& 2 (Localization)} & \textbf{Stage 3 (Context Extension)} \\
\midrule
Context Length & 8,192 & 65,536 \\
Global Batch Size (GBS) & 2,048 & 256 \\
Micro Batch Size (MBS) & 4 & 4 \\
Tokens per Batch & \texttildelow16.7M & \texttildelow16.7M \\
Learning Rate & 1e-4 & 1e-4 \\
MoE aux loss coeff & 0.002 & 0.002 \\
Training Iterations & 4,700 & 140\\
Precision & MXFP8 Mixed & MXFP8 Mixed \\
TP / PP / EP & 1 / 2 / 4 & 1 / 2 / 4 \\
Compute Infrastructure & AWS P6 (B300) & AWS P6 (B300) \\
\bottomrule
\end{tabular}
\end{table}

\subsubsection{Supervised Fine-Tuning (SFT) }
Instruction tuning approaches have demonstrated significant improvements in alignment and instruction-following behavior relative to pretrained-only models \citep{ouyang2022instructgpt}. Our post-training methodology scales supervised fine-tuning (SFT) to improve reasoning performance, conversational robustness, and instruction-following accuracy across diverse task settings. The SFT phase was initialized from the localized checkpoint produced by our CPT pipeline, rather than the original Nemotron-3-Nano-30B-A3B-Base-BF16 weights, ensuring that Hebrew linguistic knowledge acquired during pretraining is preserved throughout alignment. 

The SFT corpus consists of high-fidelity conversational and reasoning-oriented datasets that span seven task categories, with a total volume of \textbf{2M samples}, designed to balance task diversity while maintaining stable optimization dynamics. We employ two complementary paradigms for data construction: \textit{localized knowledge distillation}, which transfers reasoning-rich supervision from English-centric corpora into Hebrew through structure-preserving translation, and \textit{synthetic constraint-driven generation}, which produces linguistically controlled samples targeting instruction adherence and structured reasoning behaviors, inspired by Self-Instruct \citep{wang2022selfinstruct}. We adopted the standard \textbf{Nemotron-3-Nano chat template} to maintain compatibility with the foundation model's instruction-following logic and ensure consistent prompt formatting across Hebrew and English benchmarks. The complete dataset composition and sample breakdown are described in Section~\ref{sec:sft-training}.

\subsubsection*{Training Hyperparameters}

The primary hyperparameters and performance metrics for the SFT phase are summarized in Table~\ref{tab:sft_hyper}.

\begin{table}[H]
\centering
\caption{SFT Infrastructure and Training Hyperparameters}
\label{tab:sft_hyper}
\small
\begin{tabular}{ll}
\toprule
\textbf{Hyperparameter / Infrastructure} & \textbf{Value} \\
\midrule
Compute Infrastructure & 4x AWS P6 (NVIDIA B300 Blackwell) \\
Interconnect & AWS Elastic Fabric Adapter (EFA) \\
Average Step Time & 10.75 seconds \\
GPU Tensor Core Utilization & \texttildelow99\% (Saturated) \\
Max Sequence Length & 65,536 \\
Training Steps & 1150 \\
Global Batch Size (GBS) & 256 \\
Micro Batch Size (MBS) & 8 \\
Precision & FP8/BF16 (MXFP8 Mixed) \\
Tensor Parallel (TP) & 4 \\
Pipeline Parallel (PP) & 2 \\
Expert Parallel (EP) & 2 \\
Optimizer & AdamW \\
\bottomrule
\end{tabular}
\end{table}

\subsection{Distributed Infrastructure}
\label{sec:distributed_infrastructure}

The computational feasibility of our large-scale localization effort depends critically on recent advances in training efficiency. FP8 mixed-precision training, introduced by \citet{micikevicius2022fp8} and implemented in the NVIDIA Transformer Engine, enables substantially higher throughput than BF16 or FP16 while maintaining numerical stability in transformer training. We leverage MXFP8 precision throughout both the CPT and SFT phases. Packed sequence training, as described by \citep{krell2021efficient}, eliminates padding inefficiencies by packing multiple variable-length sequences into fixed-length batches, significantly improving hardware utilization during SFT. The Megatron-LM framework \citep{shoeybi2019megatron} provides the distributed training backbone, combining pipeline and expert parallelism to balance communication and compute overhead, while ZeRO optimization \citep{rajbhandari2020zero} enables efficient memory scaling across the data, pipeline, and expert parallelism dimensions required by our architecture. Prior work by \citet{fedus2021switch} and \citet{jiang2024mixtral} establishes sparse MoE architectures as a compute-efficient pathway to scaling model capacity, directly motivating our choice of the Nemotron-3-Nano-30B-A3B-Base-BF16 model.

We began training on an NVIDIA H200-based HyperPod cluster before transitioning to NVIDIA B300 systems. On H200 (64~GPUs), we observed a per-GPU throughput of approximately 2.8K tokens/sec, corresponding to roughly 178K tokens/sec cluster-wide. In contrast, a single B300 node (8~GPUs) achieved approximately 11.6K tokens/sec per GPU, or about 93K tokens/sec per node. These values are computed by dividing a fixed workload of 1M tokens per training step by the measured step time (5.6s for H200 HyperPod and 10.75s for B300), and normalizing by the number of GPUs.

Deploying Blackwell-generation GPUs via AWS EC2 P6 instances introduced a qualitatively new dimension to our training setup. Each B300 GPU provides approximately 280~GB of VRAM, compared to roughly 140~GB on H200 systems. As prior analyses show that transformer workloads are frequently memory-bound rather than compute-bound \citep{narayanan2021efficient}, this additional capacity directly enables two key improvements. First, it allows the micro-batch size (MBS) to be increased from 4 to 8 during SFT, driving GPU tensor core utilization from approximately 65\% to 99\% and transitioning the workload into a compute-bound regime. Second, it makes 65,536-token context lengths during later CPT stages computationally tractable without requiring aggressive reductions in batch size, directly resolving the memory-compute tradeoff characterized by \citet{narayanan2021efficient}, whose analysis of pipeline parallelism also informs our 2-way pipeline parallel configuration.

Projecting these measurements to a 100B-token training workload, the H200 cluster sustains approximately 15.4B tokens per day ($178\text{K} \times 86{,}400$ seconds), yielding a runtime of approximately 6.5 days and an estimated cost of \$52K at roughly \$8K/day. The B300 configuration processes approximately 8.0B tokens per day ($93\text{K} \times 86{,}400$ seconds), requiring about 12.5 days and costing approximately \$26.8K at roughly \$2.15K/day. Although the H200 setup provides higher absolute throughput, the B300 delivers nearly $2\times$ higher cost efficiency (tokens/sec per dollar), primarily due to improved memory-driven batch scaling, higher sustained utilization, and reduced distributed systems overhead. Collectively, these improvements enabled the execution of a 154B-token pretraining run and a 2.8B-token SFT phase at quality levels that would previously have required substantially larger compute budgets.

\subsubsection{Continuous Pre-training (CPT)}
\label{sec:cpt}

The CPT phase utilized the Megatron-Bridge training stack, optimized for hardware utilization on Blackwell-generation accelerators, and was structured into three sequential stages to stabilize the model's attention mechanisms while gradually increasing complexity.

Training was executed on AWS EC2 P6 instances utilizing NVIDIA B300 (Blackwell) GPUs interconnected via AWS Elastic Fabric Adapter (EFA), leveraging the Scalable Reliable Datagram (SRD) protocol. To manage the 30B-parameter MoE architecture built upon the Nemotron-3-Nano-30B-A3B-Base-BF16 foundation model \citep{nvidia2025nemotron}, the following parallelism configuration was implemented:

\begin{itemize}
    \item \textbf{Pipeline Parallelism (PP):} A 2-way split distributed the model layers across the Blackwell VRAM.
    \item \textbf{Expert Parallelism (EP):} A 4-way split was applied to the sparse MoE layers, optimizing expert utilization across the cluster.
    \item \textbf{Tensor Parallelism (TP):} Maintained at 1 to minimize inter-GPU communication overhead during the dense layers of the pre-training phase.
\end{itemize}

We also evaluated alternative distributed training strategies. In particular, we compared a Hugging Face Transformers + DeepSpeed ZeRO-3 (HF+DS-Z3) setup against an NVIDIA NeMo FP8-based pipeline integrated with the Megatron Bridge. The NeMo FP8 + Megatron Bridge configuration demonstrated approximately $2.2\times$ higher training throughput, consistent with prior observations that tighter integration between precision formats, parallelism strategies, and system-level optimizations can improve hardware utilization in large-scale MoE training.

\subsubsection{Supervised Fine-Tuning (SFT)}
\label{sec:sft}

Instruction fine-tuning (SFT) was conducted using the Megatron-Bridge framework, initialized from a localized checkpoint derived from the Nemotron-3-Nano-30B-A3B-Base-BF16 model \citep{nvidia2025nemotron} via the preceding CPT phase. The optimization followed a Warmup-Stable-Decay learning rate schedule with a peak learning rate of $5 \times 10^{-5}$ and 800 warmup iterations. Packed sequence training \citep{krell2021efficient} was employed to minimize padding inefficiency, and FP8 mixed-precision (\texttt{bf16\_with\_mxfp8\_mixed}) was used throughout to maximize throughput \citep{micikevicius2022fp8}.

The alignment phase was executed on 4 AWS P6 nodes, each equipped with NVIDIA B300 GPUs interconnected via AWS EFA using the SRD protocol. Moving to a Blackwell-based stack allowed for a transition from legacy multi-node HyperPod clusters to a more streamlined configuration, reducing the distributed failure surface by minimizing synchronization points.

\subsection{Evaluation}
\subsubsection{CPT Model Evaluations}
To assess the linguistic and factual knowledge acquired during the CPT phase, we evaluate our base model checkpoint on the Hebrew LLM Leaderboard \citep{shmidman2024dictalm2}, a publicly available evaluation suite designed for few-shot assessment of Hebrew base models. Evaluation covers six tasks: SNLI, QA,  Sentiment Classification, Winograd, Translation, and Israeli Trivia. Our Model is compared against Gemma-3-27B \citep{gemma2025}, DictaLM-3.0-24B-Base \citep{shmidman2026dictalm3}, and the Nemotron-3-Nano-30B-A3B-Base-BF16 foundation model prior to CPT included to directly quantify the localization gains introduced by our training pipeline.

For English reasoning capabilities, we evaluated performance on a set of established benchmarks, including HellaSwag, GSM8K, and a psychometric evaluation (Psi). The psychometric test is designed to assess higher-order cognitive abilities such as logical consistency, pattern recognition, and abstract reasoning, providing a complementary perspective to standard NLP benchmarks by approximating structured reasoning tasks. Our model achieves competitive performance on this evaluation, closely matching the pretrained baseline, indicating that core reasoning capabilities were largely preserved during adaptation. Additionally, we conducted evaluations on both HellaSwag and GSM8K, without task-specific fine-tuning. While performance on these benchmarks decreased relative to the pretrained model, the results highlight the expected trade-off between domain specialization and general English reasoning.

\subsubsection{SFT Model Evaluations}
\label{sec:SFT Model Evaluations}

Comprehensive evaluation across diverse benchmarks follows the holistic assessment paradigm proposed by HELM \citep{liang2022helm} to ensure stability and reasoning transparency. All evaluations were conducted in a zero-shot setting.

\subsubsection*{Evaluation Benchmarks and Methodology}

To provide a high-fidelity assessment of localized capabilities, expert linguists were employed to localize and translate established English-centric reasoning benchmarks into high-register Hebrew. Similar human-mediated localization practices have been shown to improve evaluation validity in multilingual benchmarking settings \citep{leScao2022bloom, ustun2024aya}.

\paragraph{Choice of Plausible Alternatives (COPA).} COPA evaluates a system's ability to perform open-domain commonsense causal reasoning \citep{roemmele2011copa}. Each question provides a premise and two plausible alternatives, requiring the model to identify the more likely cause or effect.

\paragraph{AI2 Reasoning Challenge (ARC).} The ARC benchmark consists of elementary and middle-school science questions designed to evaluate scientific knowledge and multi-step reasoning ability \citep{clark2018arc}. ARC questions are intentionally constructed to resist shallow pattern matching and require fact integration across multiple knowledge sources.

\paragraph{HellaSwag.} HellaSwag evaluates grounded commonsense inference by requiring models to select the most plausible continuation of a scenario \citep{zellers2019hellaswag}. The dataset is adversarially filtered to remain easy for humans while challenging for language models.

\paragraph{Massive Multitask Language Understanding (MMLU).} MMLU measures multitask accuracy across 57 academic subjects spanning STEM, humanities, social sciences, and professional domains \citep{hendrycks2021mmlu}. Strong performance requires extensive world knowledge combined with advanced reasoning and problem-solving capabilities.

\paragraph{Grade School Math 8K (GSM8K).} GSM8K is a dataset of approximately 8,500 grade-school mathematical word problems requiring multi-step reasoning and intermediate calculations \citep{cobbe2021gsm8k}. Problems typically require between two and eight reasoning steps.

\paragraph{Psychometric Entrance Test (Psi).} The Psychometric Entrance Test (Psi), produced by the National Institute for Testing and Evaluation (NITE), is used for higher-education admissions in Israel and evaluates verbal reasoning, quantitative reasoning, and English proficiency \citep{nite2023psi}. The model was tested on both localized Hebrew sections and the original English components to verify cross-lingual reasoning stability.

\subsubsection{Human Preference Arena Evaluation}
To complement automated benchmark performance with direct human preference signal, we conducted a cross-model preference arena comparing our model after SFT phase against two external baselines: \texttt{google/gemma-3-27b-it} and \texttt{DictaLM-3.0-24B-Thinking}. Automated benchmarks are known to inadequately capture the response quality dimensions that matter most in deployment — fluency, helpfulness, and cultural appropriateness — particularly for non-English languages where evaluation instruments are scarce (Liang et al., 2022; Scao et al., 2022). The arena provides a direct, distribution-free assessment of human preference under realistic usage conditions.

\subsubsection*{Methodology}
Annotators were presented with pairs of model responses to identical prompts drawn from naturalistic Hebrew instruction-following tasks. Each pair was evaluated under a blinded, pairwise forced-choice protocol: annotators selected the preferred response or indicated a tie. Responses were assessed holistically, with per-dimension preference labels recorded across four criteria: Relevance, Completeness, Hallucination/Factuality, and Language Quality. All model identities were hidden throughout. The arena was run as a round-robin tournament, yielding a fully connected comparison graph across all three model pairs.
Statistical significance was assessed using a Bradley-Terry-Luce Cumulative Link Mixed Model (CLMM) with random annotator effects. Multiple comparisons were corrected using the Holm-Bonferroni procedure. A pair is reported as significant only if it survives Holm correction.

\section{Results}
\subsection{CPT Model}
Hebrew and English base model performance following CPT, including comparisons against state-of-the-art base models such as DictaLM-3.0-24B \citep{shmidman2026dictalm3} and Gemma-3-27B \citep{gemma2025}, is reported in Table~\ref{tab:merged_base_eval}.
\begin{table}[H]
\centering
\caption{Hebrew and English Base Model Performance Comparison with Official Benchmarks}
\label{tab:merged_base_eval}
\scriptsize
\begin{tabular}{llcccc}
\toprule
\textbf{Benchmark} & \textbf{Lang} & \textbf{Nemotron-3-Nano} & \textbf{Our Model} & \textbf{Gemma-3-27B} & \textbf{DictaLM-3.0} \\
\midrule
\multicolumn{6}{l}{\textit{Hebrew NLP Benchmarks}} \\
SNLI Accuracy & HE & 88.81 & 91.2 & 85.2 & 86.0 \\
QA TLNLS& HE & 71.09 & 73.2 & 78.3 & 76.6 \\
Sentiment Acc& HE & 68.4 & 63.5 & 65.0 & 68.7 \\
Winograd& HE & 74.1 & 74.1 & 81.7 & 83.8 \\
Translation BLEU & HE & 33.1 & 33.6 & 36.5 & 37.2 \\
Israeli Trivia & HE & 58.14 & 72.1 & 70.4 & 82.7 \\
\midrule
\textbf{Hebrew Average} & & \textbf{65.61} & \textbf{68.0} & \textbf{69.5} & \textbf{72.5} \\
\midrule
\multicolumn{6}{l}{\textit{English Reasoning}} \\
HellaSwag (EN) & EN & \textbf{85.6} & 65.6 & \textbf{85.6} & 56.9 \\
GSM8K (EN) & EN & \textbf{92.3} & 83.5 & \textbf{82.6} & 84.1 \\
Psychometric Psi (EN) & EN & 88.7 & 88.2 & 94.7 & 89.7 \\
\midrule
\textbf{English Average} & & \textbf{88.9} & \textbf{79.1} & \textbf{87.6} & \textbf{76.9} \\
\bottomrule
\end{tabular}
\end{table}

\subsection{SFT Model}
\subsubsection*{Comparative Performance}

The performance of Our Model was rigorously compared against state-of-the-art open-source and proprietary baselines, including its pre-CPT version (the Nemotron-3-Nano-30B-A3B-Base-BF16), Gemma-3-27B-IT \citep{gemma2025}, and DictaLM-3.0-24B-Thinking \citep{shmidman2026dictalm3}. Comparative results demonstrate specialized proficiency in localized Hebrew reasoning tasks without measurable degradation in general English reasoning performance, consistent with findings from multilingual adaptation studies \citep{conneau2020xlmr, gururangan2020dont}. Full post-training evaluation results for the SFT stage are reported in Table~\ref{tab:post_eval}.

\begin{table}[H]
\centering
\caption{Comparative Performance across Hebrew and English Reasoning Benchmarks (Accuracy \%)}
\label{tab:post_eval}
\small
\begin{tabular}{llccc}
\toprule
\textbf{Benchmark} & \textbf{Lang} & \textbf{Our Model} & \textbf{Gemma-3-27B-IT} & \textbf{DictaLM-3.0-Thinking} \\
\midrule
\multicolumn{5}{l}{\textit{Hebrew Reasoning}} \\
Copa (HE)            & HE & 91.9          & \textbf{93.3} & 88.0 \\
ARC-AI2 (HE)         & HE & 88.0          & \textbf{91.4} & 91.2 \\
HellaSwag (HE)       & HE & 58.9          & \textbf{63.6} & 61.7 \\
MMLU (HE)            & HE & 68.4          & \textbf{72.5} & 60.2 \\
GSM8K (HE)           & HE & \textbf{83.3} & 82.8          & 70.2 \\
Psychometric Psi (HE)& HE & 52.5          & \textbf{54.3} & 42.3 \\
\midrule
\textbf{Hebrew Average} & & \textbf{73.8} & \textbf{76.3} & \textbf{68.9} \\
\midrule
\multicolumn{5}{l}{\textit{English Reasoning}} \\
HellaSwag (EN)        & EN & 82.5          & 89.8          & \textbf{91.1} \\
GSM8K (EN)            & EN & 83.8          & \textbf{91.7} & 86.1 \\
Psychometric Psi (EN) & EN & 91.6          & 94.5          & \textbf{95.3} \\
\midrule
\textbf{English Average} & & \textbf{86.0} & \textbf{92.0} & \textbf{90.8} \\
\bottomrule
\end{tabular}
\end{table}

\subsection{Human Preference Arena}
\begin{table}[H]
\centering
\caption{Phase 2 Arena --- Overall Preference Results}
\label{tab:arena_overall}
\begin{tabular}{lcccc}
\toprule
\textbf{Model} & \textbf{Battles} & \textbf{Wins} & \textbf{Ties} & \textbf{Losses} \\
\midrule
\texttt{google/gemma-3-27b-it}    & 200 & \textbf{122 (61.0\%)} & \textbf{40 (20.0\%)} & \textbf{38 (19.0\%)} \\
Our Model                          & 197 & 77 (39.1\%)           & 39 (19.8\%)          & 81 (41.1\%) \\
\texttt{DictaLM-3.0-24B-Thinking} & 183 & 41 (22.4\%)           & 21 (11.5\%)          & 121 (66.1\%) \\
\bottomrule
\end{tabular}
\end{table}
The ranking by every metric is \textbf{Gemma-3-27B-IT $>$ Our Model $>$ DictaLM-3.0-24B-Thinking},
with all three pairs Holm-significant (see Table~\ref{tab:arena_h2h}).

\begin{table}[h]
\centering
\caption{Phase 2 Arena --- Direct Head-to-Head Results}
\label{tab:arena_h2h}
\resizebox{\textwidth}{!}{%
\begin{tabular}{lccccr}
\toprule
\textbf{Matchup} & \textbf{1st place} & \textbf{2nd place} & \textbf{Ties} & \textbf{n} & \textbf{Decisive votes} \\
\midrule
Our Model vs. \texttt{gemma-3-27b-it} & 22 & \textbf{56} & 29 & 107 & \textbf{28.2\% / 71.8\%} \\
Our Model vs. \texttt{DictaLM-3.0-24B-Thinking} & \textbf{55} & 42 & 10 & 90 & \textbf{68.8\% / 31.2\%} \\
\texttt{gemma-3-27b-it} vs. \texttt{DictaLM-3.0-24B-Thinking} & \textbf{66} & 16 & 11 & 93 & \textbf{80.5\% / 19.5\%} \\
\bottomrule
\end{tabular}%
}
\end{table}

\subsection{Inference Speed}

Inference throughput was benchmarked on a single NVIDIA RTX 6000 PRO GPU 
under identical hyperparameters and context length across all three models. 
Our model, activating approximately 3B parameters per forward pass~\citep{nvidia2025nemotron}, achieved 
roughly $9\times$ higher token throughput compared to both 
\texttt{gemma-3-27b-it} and \texttt{DictaLM-3.0-24B-Thinking}, each 
activating approximately 27B and 23B parameters respectively. This result 
is consistent with the active-parameter ratio inherent to the sparse MoE 
architecture~\citep{fedus2021switch, jiang2024mixtral}, and confirms that 
the efficiency advantage is realized end-to-end under realistic serving 
conditions.

\section{Discussion}

\subsection*{Continuous Pre-training}
Including the Nemotron-3-Nano-30B-A3B-Base-BF16 pre-CPT checkpoint as a baseline provides a direct 
measurement of the gains attributable to the CPT phase alone. Our model improves upon this 
baseline by 2.39 percentage points on average. The most pronounced gain is on Israeli Trivia 
(+13.96 points), confirming that culturally grounded Hebrew world knowledge --- systematically 
underrepresented in the foundation model's English-dominant corpus --- is effectively injected 
through the localized pretraining pipeline. SNLI Accuracy improves meaningfully (+2.39 points), 
reflecting stronger abstract semantic reasoning over Hebrew text. More modest gains on Sentiment 
and Winograd are consistent with prior multilingual adaptation findings: tasks sensitive to 
morphological surface form require dedicated post-training alignment to fully materialize 
\citep{pfeiffer-etal-2020-mad, gururangan2020dont}.

Comparing against external baselines at the base model stage, our model achieves the highest 
SNLI Accuracy (91.2\%) across all evaluated models and outperforms Gemma-3-27B on Israeli 
Trivia (72.1\% vs.\ 70.4\%), demonstrating effective injection of culturally grounded Hebrew 
world knowledge. DictaLM-3.0-24B-Base leads on four of six tasks and achieves the highest 
base-model Hebrew average (72.5\%), reflecting its fully Hebrew-centric training regime. 
However, as shown in Table~\ref{tab:post_eval}, instruction tuning substantially closes 
this gap.

\subsection*{Supervised Fine-Tuning}
After SFT, our model achieves a Hebrew average of \textbf{73.8\%}, surpassing 
DictaLM-3.0-24B-Thinking by 4.9 percentage points (68.9\%) and narrowing the gap with 
Gemma-3-27B-IT (76.3\%) to 2.5 points --- a strong result given the approximately 9$\times$ 
difference in active parameters at inference time. Task-level strengths are consistent across 
the evaluation suite. Our model leads on Copa~(HE) at 91.9\%, reflecting robust commonsense 
causal reasoning in Hebrew, and achieves 88.0\% on ARC-AI2~(HE), indicating reliable 
integration of factual knowledge under multi-step reasoning demands. On GSM8K~(HE), our model 
scores 83.3\%, marginally ahead of both DictaLM-3.0-Thinking (70.2\%) and Gemma-3-27B-IT 
(82.8\%), confirming that mathematical reasoning is fully preserved through localization. On 
MMLU~(HE), our model achieves 68.4\%, ahead of DictaLM-3.0-Thinking (60.2\%) but below 
Gemma-3-27B-IT (72.5\%), reflecting the broader world-knowledge advantage of a larger dense 
model. The Psychometric Psi~(HE) score of 52.5\% trails Gemma (54.3\%) but exceeds 
DictaLM-3.0-Thinking (42.3\%), a meaningful gap on a high-stakes structured reasoning 
benchmark. English reasoning fidelity is well maintained throughout, with an English average 
of 86.0\%, confirming that neither CPT nor SFT induces measurable catastrophic forgetting of 
general-purpose capabilities \citep{gururangan2020dont, conneau2020xlmr}.

\subsection*{Human Preference Arena}
Human preference evaluation confirms and extends the benchmark picture. Our model decisively 
outperforms DictaLM-3.0-24B-Thinking, winning 68.8\% of decisive votes across 90 battles, 
with the strongest margins on Relevance and Completeness. Against Gemma-3-27B-IT, our model 
trails on aggregate preference (28.2\% of decisive votes). This gap must be read against a 
fundamental compute asymmetry: Gemma-3-27B activates approximately 27B parameters per forward 
pass, while our MoE architecture activates approximately 3B, a roughly 9$\times$ reduction in 
active inference compute \citep{fedus2021switch, jiang2024mixtral}. On the dimensions most 
sensitive to Hebrew specialization, our model remains competitive, leading on Factuality and 
Hebrew Language Quality. These advantages are consistent with its Israeli Trivia and GSM8K 
results, and are directly attributable to the carefully curated multilingual data mixture 
across all three CPT phases, which jointly ground the model in culturally rich Hebrew knowledge 
while preserving cross-lingual reasoning fidelity \citep{gururangan2020dont, pfeiffer-etal-2020-mad}. 
All three arena pairs are statistically significant under Holm-corrected CLMM analysis, 
yielding an unambiguous ranking of Gemma-3-27B-IT $>$ Our Model $>$ DictaLM-3.0-24B-Thinking. 
Our model reaches this position at one-ninth the active inference compute of Gemma, making it 
the most compute-efficient open-weight Hebrew model at this quality level.

\subsection*{Conclusion}
Taken together, these results validate the central thesis of Hebrew-specialized MoE 
localization. Our model consistently surpasses DictaLM-3.0-24B-Thinking --- the 
previous open-weight Hebrew frontier --- across automated benchmarks and human preference 
evaluation, while remaining competitive with Gemma-3-27B-IT on factuality, mathematical 
reasoning, and cultural grounding. It achieves this at approximately one-ninth the active 
inference compute cost of Gemma, and at substantially lower fine-tuning cost, making it the 
most efficient open-weight option for Hebrew-specialized deployment. The curriculum-ordered 
CPT pipeline, anti-forgetting data design, and dedicated Hebrew alignment corpus each 
contribute measurably to this outcome, providing a reproducible blueprint for sovereign 
language model development in morphologically rich, lower-resource languages.

\bibliographystyle{plainnat}
\bibliography{references}

\end{document}